\pdfoutput=1

\documentclass[letterpaper, 10 pt, conference]{ieeeconf}

\usepackage{amsmath}
\usepackage[noend]{algpseudocode}
\usepackage{verbatim}
\usepackage{makecell}
\usepackage{graphicx}
\usepackage{multirow}
\usepackage[ruled]{algorithm2e}

\IEEEoverridecommandlockouts 
\title{\LARGE \bf
Efficient Large-Scale Face Clustering Using an Online Mixture of Gaussians
}

\author{David Montero, Naiara Aginako, Basilio Sierra and Marcos Nieto
\thanks{David Montero an Marcos Nieto are with Vicomtech Foundation, Basque Research and Technology Alliance (BRTA), Mikeletegi 57, 20009 Donostia-San Sebasti\'an (Spain). Email: dmontero@vicomtech.org, mnieto@vicomtech.org}
\thanks{Naiara Aginako and Basilio Sierra are with the University of the Basque Country. Email: naiara.aginako@ehu.eus, b.sierra@ehu.eus}%
}

\begin{document}

\maketitle
\thispagestyle{empty}
\pagestyle{empty}

\begin{abstract}
In this work, we address the problem of large-scale online face clustering: given a continuous stream of unknown faces, create a database grouping the incoming faces by their identity. The database must be updated every time a new face arrives. In addition, the solution must be efficient, accurate and scalable. For this purpose, we present an online gaussian mixture-based clustering method (OGMC). The key idea of this method is the proposal that an identity can be represented by more than just one distribution or cluster. Using feature vectors (f-vectors) extracted from the incoming faces, OGMC generates clusters that may be connected to others depending on their proximity and their robustness. Every time a cluster is updated with a new sample, its connections are also updated.
With this approach, we reduce the dependency of the clustering process on the order and the size of the incoming data and we are able to deal with complex data distributions. Experimental results show that the proposed approach outperforms state-of-the-art clustering methods on large-scale face clustering benchmarks not only in accuracy, but also in efficiency and scalability.

\end{abstract}
\section{Introduction}

In recent years, face clustering by identity has been in high demand in a variety of contexts and, in some cases, to support real-time operation of application. For instance, it is required in real-time video-surveillance applications, where there is a need to maintain an updated database of subjects within the monitored area, either to control their locations \cite{rtFaceRec,rtFaceRec1} or to re-identify a subject if necessary \cite{rtFaceRec_criminal,rtFaceRec_criminal1}. 
Another usage of face clustering in a real-time application is people flow monitoring in large infrastructures. This type of application aims to generate Performance Indicators, such as "waiting times", "process throughput", "person show-up profile", "queue length overrun" and "area occupancy" \cite{Mayer2015,JIT2017}, which need to be computed for increasingly large number of cameras, and thus demanding a higher level of computation scalability.

An important issue that arises about these types of applications is data privacy \cite{privacy1,privacy2,privacy3}. Personal information that could be used to identify the subjects should not be stored (i.e. images of their faces). Recent clustering algorithms rely on Deep Neural Networks (DNN) that can infer feature vectors (f-vectors) from the targeted face images, which correspond to abstract representations of the people appearances used for training.
Although in the last years some promising methods for face rendering from f-vectors have been released \cite{face_rendering,face_rendering1}, they require to know the feature extraction network in order to generate acceptable results \cite{face_rendering1} or to train the decoding network \cite{face_rendering}. Therefore, the individual identities can be protected by hiding and securing the embedding network. Obviously, this approach has limitations, such as the possibility of matching different people that look too similar for the trained model. Therefore, the design and training of the DNN should cover as best as possible the subtle facial appearance dissimilarities. State-of-the-art face recognition models \cite{Deng2018,facenet} aim at accomplishing that goal.

\begin{figure}[t]
    \centering
    \includegraphics[width=\linewidth]{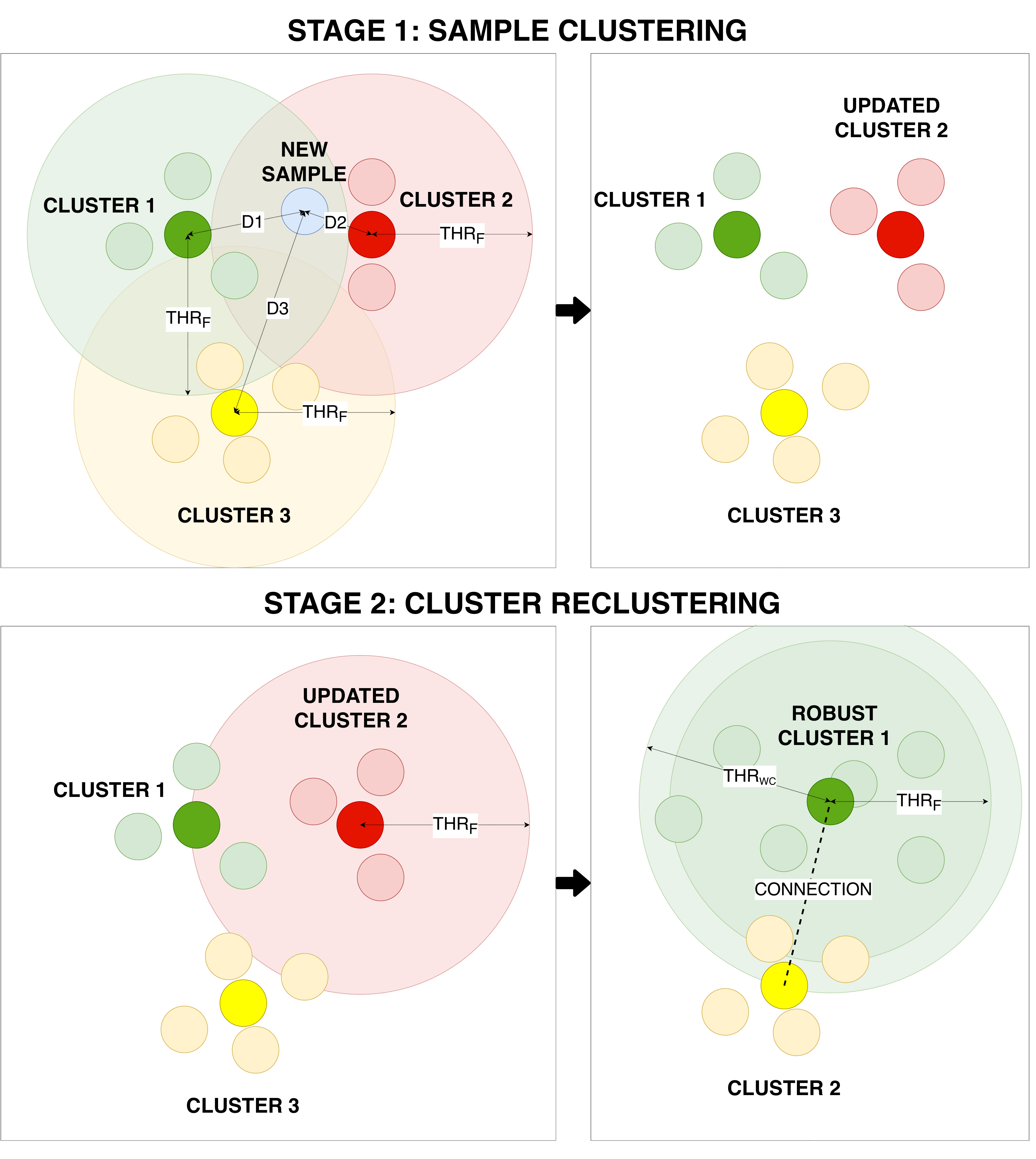}
    \caption{Example of the operation of the proposed online clustering algorithm when a new sample arrives.}
    \label{fig:clustering_example}
\end{figure}

Furthermore, these real-time applications may work in large-scale unconstrained environments, where no information about the distribution of face representations or the number of identities is available. The faces may come from multiple cameras placed in different locations and positions, so they may have different orientations, lighting conditions, partial occlusions, etc. This will lead to complex data distributions. Most traditional and state-of-the-art clustering methods are offline (\cite{GCN2019,Otto2018,KMeans,AHC}). These offline approaches are not suitable for real-time large-scale scenarios, as they need to repeat the whole clustering process every time a new sample arrives. In addition, most of them have difficulties dealing with complex data distributions.

To overcome these problems, we present an online gaussian mixture-based clustering method (OGMC). Our proposal key idea is that an identity may be represented by several distributions or clusters. Using feature vectors (f-vectors) extracted from the incoming faces, OGMC generates clusters that may be connected to others depending on their proximity and their robustness. Every time a cluster is updated with a new sample, its connections are also updated. With this approach, we reduce the dependency of the clustering process on the order and the size of the incoming data and we are able to deal with complex data distributions. The high-level idea of the method is exposed in figure \ref{fig:clustering_example}. Experimental results show that our approach outperforms state-of-the-art clustering methods on large-scale face clustering benchmarks not only in accuracy, but also in efficiency and scalability.

The rest of the paper is organized as follows. First, we present a review of the related work in section 2. Section 3 describes the proposed clustering method. In section 4 we provide the experimental results. Finally, the conclusions are given in section 5.

\section{Related Work}

\subsection{Unconstrained Face Clustering}

Face clustering in unconstrained environments has become a well-studied topic over the last few years. The addressed scenarios are increasingly complex and encompass a greater number of identities. The huge number of faces and the intra-class appearance changes that might happen due to environmental variations (e.g. pose, illumination, expression, ornaments, occlusions, resolution, image noise) lead to complex distributions of face representations.
Traditional clustering algorithms, as K-Means \cite{KMeans} or spectral clustering \cite{Spectral}, suffer from this complexity and they are not able to achieve acceptable performance in these scenarios, as they make assumptions on data distribution. For instance, K-Means tends to generate similar-sized clusters.

The new trends combine face recognition models based on DNN to extract facial f-vectors with sophisticated clustering algorithms that can group them in distinguishable identities, despite the intra-class appearance variability. Shi et al. \cite{Shi2018} proposed the Conditional Pairwise Clustering (ConPaC) algorithm, which is based on the direct estimation of an adjacency matrix using pairwise similarities between f-vectors. In \cite{GCN2019} a linkage-based face clustering algorithm is presented, where a graph convolution network decides which pairs of nodes should be linked. Lin et al. \cite{DDC} adopted an Agglomerative Hierarchical Clustering approach, considering the distance measure in the embedded space and the dissimilarity between two groups of faces. In \cite{Otto2018} an approximate rank-order clustering is presented, which predicts whether a node should be linked to its k Nearest Neighbors (kNN), and transitively merges all linked pairs.

Nevertheless, all these state-of-the-art methodologies employ offline algorithms. They process entirely the gathered data every time a new sample arrives, repeating the clustering process with increasing computational cost, in order to get an optimal result. In addition, most of them suffer from scalability problems in terms of accuracy and processing time. For instance, the complexity of ConPaC can scale up to $O(TN^3)$, where $N$ is the number of f-vectors and $T$ the number of iterations. Wang et al. \cite{GCN2019} and Otto et al. \cite{Otto2018} reduce the complexity of the proposed algorithms using kNN graphs to reduce the number of comparisons, but the computational cost is still too high to consider them for online applications.

\subsection{Online Clustering}

In recent years, numerous online clustering algorithms have emerged to tackle challenging situations. The main advantage of this type of algorithms is that they are able to process a new sample without repeating the whole clustering process. Therefore, they are the best choice when dealing with large-scale real-time scenarios, but with the added challenge of controlling and defining the learning rate (i.e. how new data updates the learnt models).

These types of algorithms have been applied in a wide variety of contexts. For instance they have gained an increasing importance in text clustering. In \cite{socialDataClustering}, the authors proposed an online clustering method for grouping data streams from social networks by their topic, using a similarity measure computed taking into account both the cluster age and the employed terms. Yin and Wang \cite{onlineTextClustering} presented an alternative text clustering method, assuming an unknown number of clusters, but below a maximum. Online clustering algorithms have also been employed for unsupervised representation learning \cite{Zhan2020}, where the cluster centroids evolve dynamically, keeping the classifier stably updated. Another example of online clustering application is MalFamAware \cite{onlineMalwareClustering}, an algorithm created to group new malwares into families to discern if they are novel or just a variant of a known sample. Even traditional offline methods as K-Means have their own online implementation \cite{onlineKMeans}.

Some novel clustering algorithms try to combine both online and offline approaches. Wang and Imura \cite{Wang2019} presented a gaussian process-based incremental neural network, where the new samples are treated as nodes, which may be connected to others. When a new sample is clustered, only the connections of its neighbours are re-evaluated. Meanwhile, in \cite{clusteringDataStreams}, samples are grouped in spherical static micro-clusters, connected together to create dynamically shaped macro-clusters. These approaches aim to achieve the same accuracy as offline methods but with an important decrease in the computation time.

\begin{figure*}
    \centering
    \includegraphics[width=\textwidth]{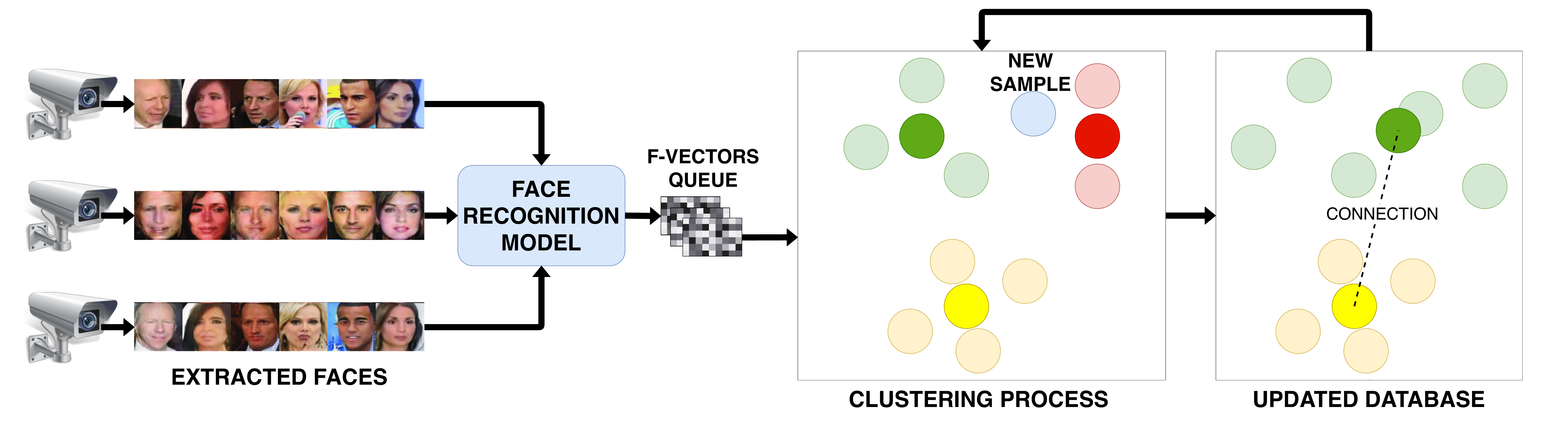}
    \caption{Example of an online clustering system. A continuous stream of face images, extracted from a set of video-surveillance cameras, are processed by a face recognition model and the extracted f-vectors are enqueued. The online clustering process updates the database with every new sample without repeating the whole process.}
    \label{fig:system_overview}
\end{figure*}

In face clustering there are also some online approaches. In \cite{onlineFaceClustering}, the authors propose an online algorithm for clustering faces in long videos. They process the data sequentially in short segments of variable length and create clusters using face representations and several spatio-temporal constraints. Nevertheless its reliance on these constraints makes their method unsuitable for unconstrained environments and highly susceptible to the order of the incoming data. Tapaswi et al. \cite{onlineFaceClustering1} presented another online face clustering method for long videos. The algorithm creates spherical clusters with a shared radius which may vary dynamically. They assume that each identity is represented by an identical distribution. Thus, this method is not able to deal with complex data distributions and, therefore, is not a valid solution for unconstrained environments.

Our proposed method aims to cover the need of an online face clustering algorithm capable of working in large-scale unconstrained environments in real time, achieving state-of-the-art results.

\section{Proposed Method} \label{sec:proposed_method}

\subsection{Problem Definition}

We consider the problem of online clustering: given a continuous stream of unknown faces, create a database grouping the incoming faces by their identity. The database must be updated every time a new face arrives, so that information on existing identities is available in real time. An example of an online clustering system is shown in Figure \ref{fig:system_overview}. For time and scalability considerations, it is not viable to regenerate the whole database in every iteration, so the algorithm must cluster the new sample using the information from the existing database and update it. Therefore, the problem can be modeled as follows:

\begin{equation}
\begin{gathered}
D_{i} = F(S_{i}, D_{i-1}) \\
D_{i} = {C_i, I_i}
\end{gathered}
\label{eq:online_clustering}
\end{equation}

\noindent where $D_{i}$ is the updated database for the i-th iteration, $F$ is the clustering process repeated in each iteration, $S_{i}$ is the i-th sample, considering a sample as a normalized N-dimensional f-vector extracted from an incoming face, and $D_{i-1}$ is the resulting database from the previous iteration. The database $D$ is represented by the group of computed clusters $C$ and by the identities $I$ to which they belong. We aim at modeling $F$ maximizing the accuracy and the scalability and minimizing the iteration time.

\subsection{Expectation-Maximization Approach} \label{sec:em_approach}

The problem modeled by the equation \ref{eq:online_clustering} is similar to the one tackled by the Expectation-Maximization (EM) algorithm \cite{EM}. The EM algorithm is a well-known iterative approach to perform maximum likelihood estimation in the presence of latent variables. It has been widely used in clustering applications \cite{EM_clustering_1,EM_clustering_2,EM_clustering_3,EM_clustering_4}. The problem formulation is the following: given a statistical model generated by a set of observed data $X$, a set of missing values $Z$, and a vector of unknown parameters $\Theta$, along with a likelihood function:

\begin{equation}
L(\Theta; X, Z) = p(X, Z | \Theta)
\label{eq:emlf}
\end{equation}

\noindent and the maximum likelihood estimate (MLE) of the unknown parameters is determined by maximizing the marginal likelihood of the observed data:

\begin{equation}
L(\Theta; X) = p(X | \Theta)
\label{eq:emmle}
\end{equation}

The EM algorithm aims to find the MLE of the marginal likelihood by iteratively applying two steps:
\begin{itemize}
    \item Expectation Step: estimates the values of the missing data $Z$, using the observed data $X$ and the current estimation of the parameters $\Theta_{i}$.
\begin{equation}
Z = F(\Theta_{i}, X)
\label{eq:emes}
\end{equation}
    \item Maximization Step: update the parameters $\Theta_{i+1}$ using the observed data $X$ and the new estimated data $Z$.
\end{itemize}
\begin{equation}
\Theta_{i+1} = F(X, Z)
\label{eq:emms}
\end{equation}

In our context, we apply the EM algorithm assuming that, at each iteration, the observed data is the group of processed samples $S$, the parameters are the features of the computed clusters $C$ and the missing value we want to estimate is the identity of the cluster to which the new sample belongs $I$.

Thus, in the estimation step, using the parameters of the clusters computed with the already processed samples, the algorithm decides whether the new sample should be merged with an existing cluster or create a new one. Then, in the maximization step, the parameters of the involved cluster are updated.

\begin{figure*}
    \centering
    \includegraphics[width=\linewidth]{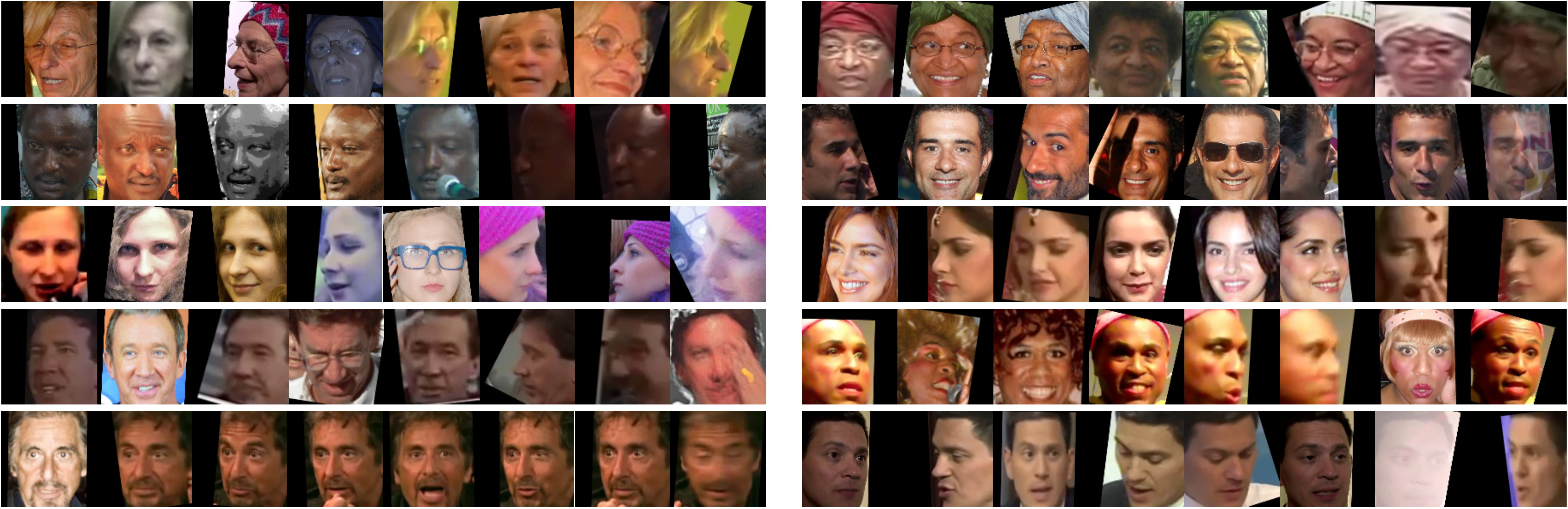}
    \caption{Examples of complex identities extracted from the IJB-C dataset. Different variations in lighting conditions, occlusions, perspective and facial attributes such as beard, glasses, hat or hair can be observed. Trying to group all faces in just one cluster may lead to errors due to overly permissive thresholds and poor quality centroids.}
    \label{fig:ijbc_face_conditions}
\end{figure*}

\begin{figure}
    \centering
    \includegraphics[width=\linewidth]{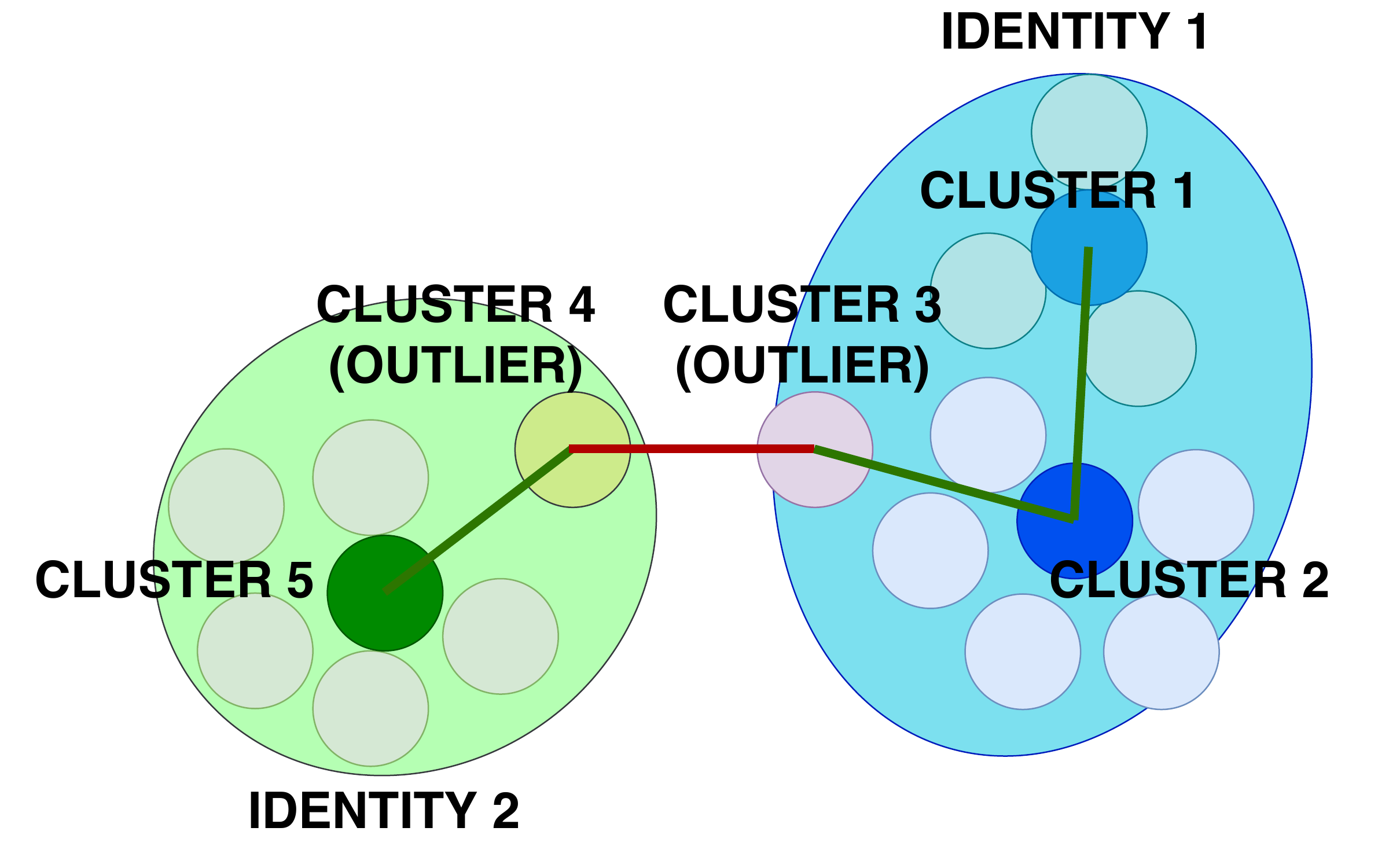}
    \caption{Example of an erroneous connection caused by outliers. Clusters 1, 2 and 5 are robust while clusters 3 and 4 are non-robust. These problems can be avoided by applying the defined connection rules.}
    \label{fig:wrong_connections}
\end{figure}

Nevertheless, our method follows a variant of this two-steps approach. Every time a cluster is updated, a new EM algorithm is launched, called cluster reclustering, where we check if the updated cluster can be fused with others. This is an iterative process, so a cluster may be fused multiple times. Thus, our algorithm is divided into two stages:

\begin{itemize}
    \item Sample clustering stage: the new sample is clustered, updating an existing cluster or creating a new one.
    \item Cluster reclustering stage: iteratively tries to fuse the updated cluster with the rest of the clusters in the database.
\end{itemize}

The two-stages process is illustrated in Figure \ref{fig:clustering_example}, and described in subsequent sections.

\subsection{Cluster Connections} \label{sec:cluster_connections}

As we want our algorithm to work in unconstrained environments, it must be able to deal with complex data distributions. Therefore, we must consider the possibility that an identity is represented by more than one cluster. For instance, facial attributes of a person may change for different reasons (i.e. glasses, beard, hair, perspective, lighting conditions, ...), and trying to group all faces in just one cluster may lead to errors due to overly permissive thresholds and poor quality centroids. Several examples of complex identities extracted from the IJB-C dataset \cite{Maze2018} are exposed in Figure \ref{fig:ijbc_face_conditions}.

For this reason, we introduce the concept of cluster connection. A connection between two clusters implies that they belong to the same identity, but they represent different distributions in the feature-space. In other words, there is a high similarity between both clusters, but this similarity is not high enough to fuse them (see Figure \ref{fig:clustering_example}). This way, the algorithm creates trees of connected clusters to deal with complex data distributions.

Furthermore, with these connections we highly reduce the dependency of our algorithm to the order of the incoming samples, as the connections of a cluster are checked and updated every time it is fused with a new sample or with another cluster.

Nevertheless, allowing multiple connections without control may lead to erroneous connections between outliers and clusters belonging to different identities (see Figure \ref{fig:wrong_connections}). To overcome this problem, we create the concept of robust clusters and several rules to control the connections. A cluster is considered robust if it is composed of, at least, $ns_r$ samples, so that we can ensure that it is not a group of outliers. The connection rules are the following:

\begin{itemize}
    \item A non-robust cluster shall have at most one connection, and it can only be connected to a robust cluster. With this rule, we avoid erroneous connections caused by outliers and redundant connections of non-robust members of an identity, solving problems as the one exposed in Figure \ref{fig:wrong_connections}.
    \item Two robust clusters can not be fused together. We consider a robust cluster as a valid distribution which represents a subset of an identity samples. Therefore, joining two robust clusters would result in a poorer representation of the identity and a loss of information.

    \item A robust cluster may have a maximum number of connections $nc_{max}$. This limitation is adopted to reduce the computation time, specially when checking if the connections of a cluster are still valid.
    \item Every time a cluster is fused or it is connected to new clusters, a process to check connections is triggered. This function checks if the connections of the updated cluster are still valid and that the maximum number of connections is not exceeded. Otherwise, the weakest connections are removed.

\end{itemize}

\subsection{Cluster Representation}

The next step in the creation of the clustering algorithm is the selection of the group of parameters that represents each cluster. These parameters must contain enough useful information about the cluster they represent in order to obtain accurate results. Furthermore, this information should be brief and concise, as the algorithm must be fast and scalable.

Considering these factors, we decided to model the clusters using multivariate normal distributions, as it only depends on two parameters: the mean vector $\mu$ and the covariance matrix $\Sigma$. This is a design choice also motivated because the EM algorithm works well with these kind of distributions \cite{EM_gaussian_1,EM_gaussian_2,EM_gaussian_3}. Therefore, the density function that models the probability that the N-dimensional sample $S_i$ belongs to cluster $j$ is:

\begin{equation}
P(S_i|C_j) \propto exp\left(-\frac{1}{2}\left( {S_i - \mu_j }^T \right) {\Sigma_j ^{-1}} \left( {S_i - \mu_j } \right)\right)
\label{eq:normal_dist}
\end{equation}

\noindent where the Mahalanobis distance \cite{mahalanobis} between $S_i$ and $\mu_j$ can be directly used to evaluate which is the closer centroid for a certain sample. Indeed, if we assume all dimensions are independent and have the same variance, we can operate on Mahalanobis distances to reduce the computational load of the algorithm, defined as follows:

\begin{equation}
D_M(S_i, \mu_j, \sigma_j) = \frac{1}{\sigma_j}dist(S_i,\mu_j)
\label{eq:mahalanobis_dist}
\end{equation}

\noindent where $dist(S, \mu)=||S-\mu||_2$.

\begin{figure}
    \centering
    \includegraphics[width=\linewidth]{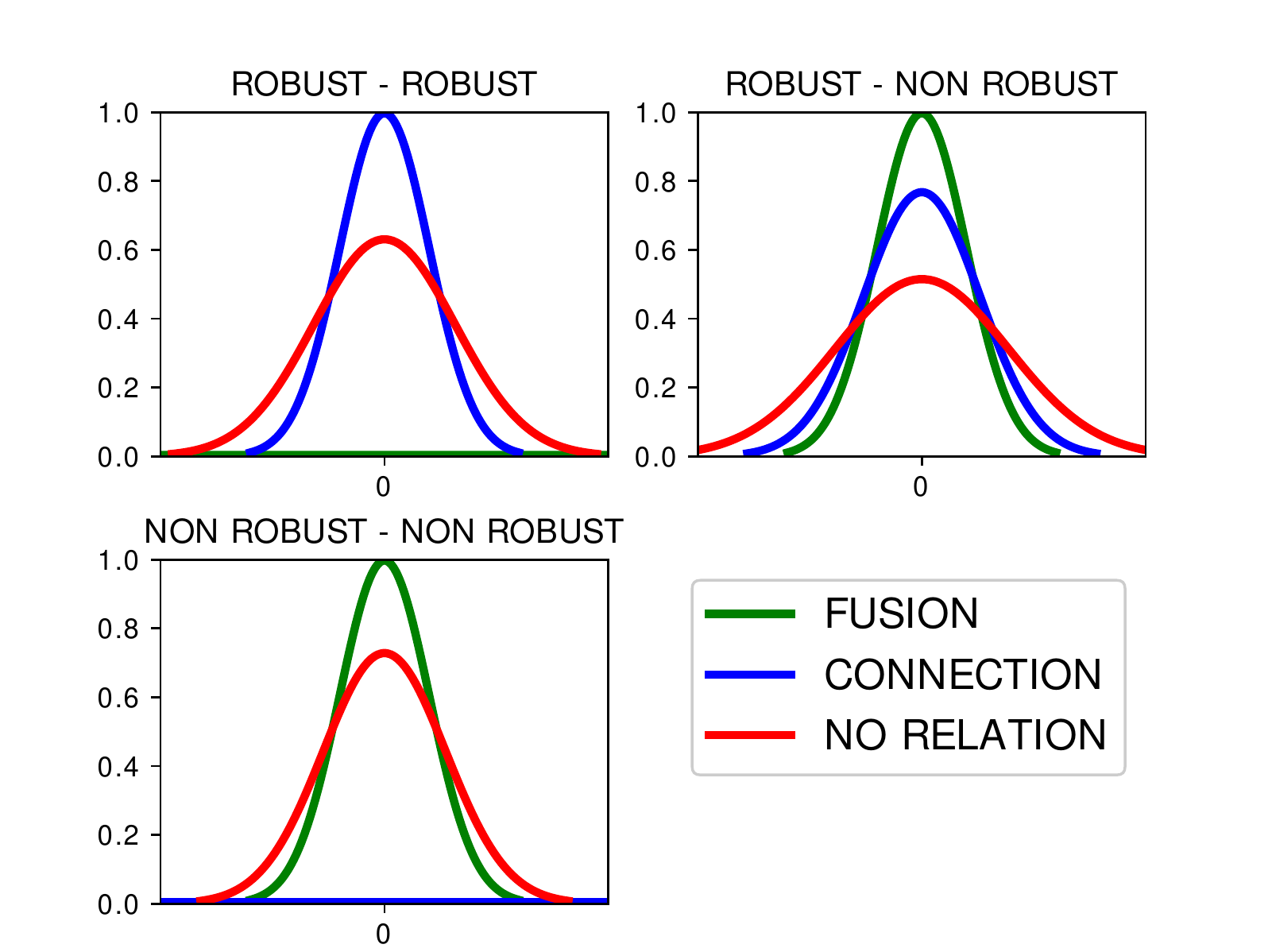}
    \caption{Normal distributions representing the probability that a cluster can be fused to, connected to or independent of another. The gaussians are centered in 0 (the minimum possible distance between the centroids). The variance depend on whether the compared clusters are robust or not and on the connection rules defined in Section \ref{sec:cluster_connections}.}
    \label{fig:cluster_distributions}
\end{figure}


The three possible cases when comparing two clusters (fusion, connection or no relation) are then modeled as a mixture of gaussians of these normal distributions, and effectively computed using the Mahalanobis distance. These distributions have all zero mean, which is equal to the minimum possible distance between two centroids. Their deviations $\sigma$ take fixed values depending on whether the compared clusters are robust or not and on the previously defined connection rules (see Figure \ref{fig:cluster_distributions}).

Therefore, we define three euclidean distance thresholds to cover all the possible cases generated by the different $\sigma$:
\begin{itemize}
    \item Fusion threshold $thr_f$: to decide whether two clusters should be fused together.
    \item Weak connection threshold $thr_{wc}$: to connect a robust cluster with a non-robust one.
    \item Strong connection threshold $thr_{sc}$: to connect two robust clusters.
\end{itemize}

The mean of a cluster is represented by its centroid $C^*$, computed by normalizing the sum of the features of all the samples contained in the cluster:

\begin{equation}
\begin{gathered}
SC_{j} = \sum S \in cluster_{j} \\
C_{j}^* = \frac{SC_{j}}{||SC_{j}||_2}
\end{gathered}
\label{eq:centroid_comp}
\end{equation}

Thus, the following information is stored for each cluster:
\begin{itemize}
    \item N-dimensional centroid ($C^*$)
    \item Sum of the belonging samples features ($SC$)
    \item Number of belonging samples ($ns$)
    \item Index of the belonging samples ($sIdx$)
    \item Index of the connected clusters ($cIdx$)
    \item Distance to the connected clusters ($cDist$)
\end{itemize}

\subsection{Method Implementation}

\begin{figure}
    \centering
    \includegraphics[width=\linewidth]{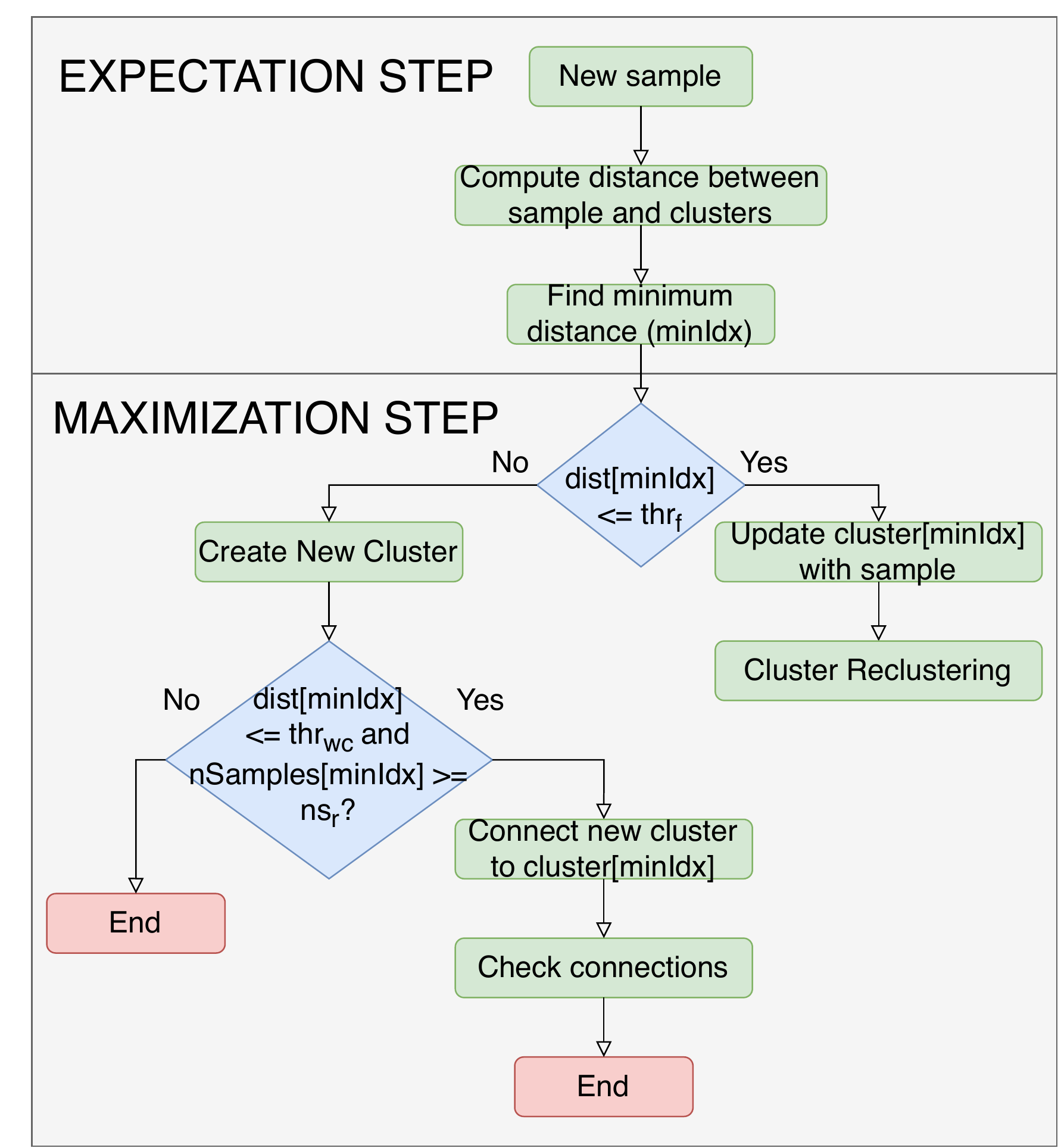}
    \caption{Diagram describing the sample clustering stage of the proposed clustering algorithm.}
    \label{fig:cluster_diagram_p1}
\end{figure}

As described in section \ref{sec:em_approach}, our clustering method is divided into two stages. The first one, the sample clustering stage, is illustrated in Figure \ref{fig:cluster_diagram_p1}. In this stage, the new incoming sample is processed. The first step is to compute the distance between the normalized sample vector and the normalized centroids of the existing clusters in the database. Different distances may be considered, but we selected the euclidean distance ($dist$). We made this decision because the employed face recognition model was trained using the cosine similarity \cite{Deng2018} and, for normalized vectors, it is inversely proportional to the euclidean distance:
\begin{equation}
dist(V1, V2) = \sqrt{2(1 - cosSim(V1, V2))}
\label{eq:euclidean_cossim}
\end{equation}

The distances are computed in parallel taking advantage of the capabilities of a GPU architecture. This way, we reduce the impact of the clusters number in the processing time. For this reason, the normalized centroids are stored directly in the GPU memory.

Once the distances have been computed, they are copied to the RAM memory and the minimum distance is selected using the CPU. If this distance is less than $thr_f$, the sample is fused with the selected cluster. Otherwise, it is used to create a new cluster. If a new cluster is created, the algorithm checks if it can be connected to the minimum distance cluster. This connection happens if the selected cluster is robust and if the distance is not higher than $thr_{wc}$. If there is a connection, the algorithm checks if the number of connections of the robust cluster has exceeded the maximum permitted ($nc_r$), and, if necessary, erase the weakest connections.

\begin{figure}
    \centering
    \includegraphics[width=\linewidth]{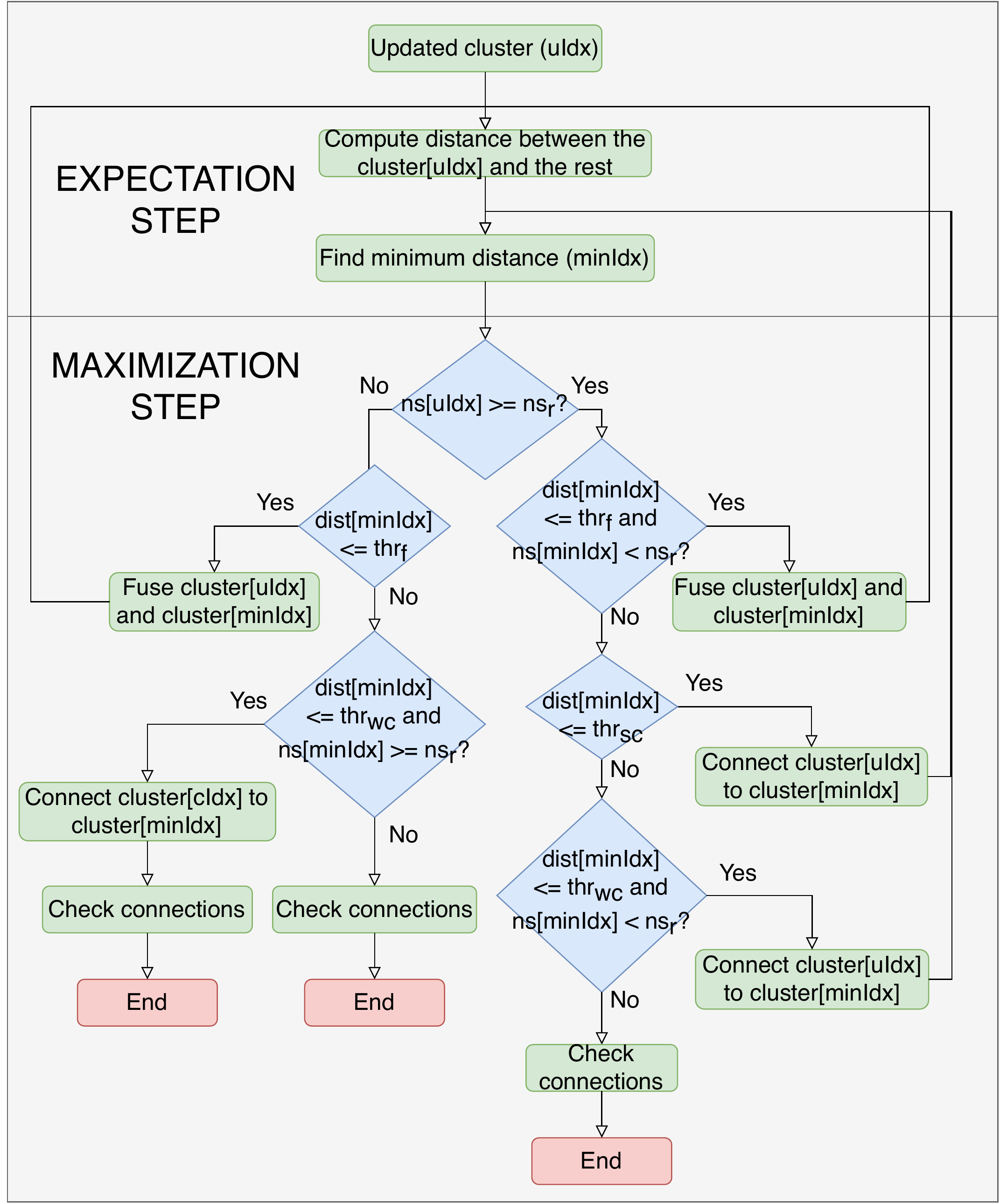}
    \caption{Diagram describing the cluster reclustering stage of the proposed clustering algorithm.}
    \label{fig:cluster_diagram_p2}
\end{figure}

If the new sample $S_i$ is used to update an existing cluster, the cluster reclustering stage is launched (see Figure \ref{fig:cluster_diagram_p2}). This stage is composed of an iterative EM algorithm where the updated cluster is tried to be fused or connected with the rest of the clusters in the database, using the parameters $thr_f$, $thr_{wc}$, $thr_{sc}$ and $ns_r$, and the connection rules defined in section \ref{sec:cluster_connections}.

The selection of the minimum distance is computed in the same way as in the previous stage. If the updated cluster is fused with another one, the distances are computed again for the new updated cluster and the process is repeated. If the updated cluster is connected to another and the updated cluster is robust, the algorithm searches for the next minimum distance and repeats the maximization step.

Before the cluster reclustering stage ends, the algorithm checks if the connections of the updated cluster are still valid, as the centroid of cluster may have changed due to a fusion. Finally, it checks that the number of connection of the updated cluster and of the connected clusters do not exceed the maximum number of connections allowed.

The results of the clustering process are extracted after each iteration using a simple recursive function and the variables $sIdx$ and $cIdx$ to merge the samples of all connected clusters.


\section{Experiments}

A series of experiments have been conducted to demonstrate the potential of the proposed clustering algorithm OGMC. The experiments are divided into two groups. The first group of aims to measure the performance of OGMC in accuracy and processing time, comparing it with other traditional and state-of-the-art offline clustering methods.

The second group focuses on testing its scalability, measuring the drop in accuracy and the increase in processing time as the number of data samples grows.

Furthermore, an ablation study is presented in order to validate the design decisions, compare the contribution of the different parts of the algorithm and measure the degree of dependency of the model parameters on the face recognition network and the train and test datasets.

Finally, to demonstrate the effectiveness of OGMC beyond face recognition, an additional experiment is conducted with DeepFashion \cite{deepfashion}, a well-known dataset used for clothes retrieval.

The server used to carry out the experiments was equipped with an NVIDIA Tesla V100 GPU and an Intel Xeon Gold 6230 CPU.


\subsection{Parameter Tuning}

As discussed in section \ref{sec:proposed_method}, the OGMC depends only on 5 parameters: three distance thresholds ($thr_f$, $thr_{wc}$, $thr_{sc}$), the minimum number of samples to classify a cluster as robust ($ns_r$) and the maximum number of connections allowed for a robust cluster ($nc_r$). This number of parameters is relatively low taking into account that it is an online method which can operate with database regardless of its magnitude, and compared to other state-of-the-art clustering methods as, for example, GCN \cite{GCN2019}, which requires to train a DNN in addition to three parameters. Furthermore, their values are bounded within certain limits and they are easy to adjust, as it is explained below.

The distance thresholds are floating-point numbers bounded between 0.0 and 2.0, as they represent the euclidean distance between two normalized vectors. These are by far the most sensitive parameters of the algorithm, as small modifications of their values have significant impact on the output.

The minimum number of samples to classify a cluster as robust ($ns_r$) is an integer equal to or greater than 1. On the one hand, the lower its value, the higher number of robust clusters, which may be undesiderable because of errors in connections between outlayers of different identities and an increase in the processing time due to circular connections. On the other hand, if $ns_r$ is too high, the number of robust clusters may not be enough to connect all the non-robust clusters together. We have empirically found that a range between 3 and 6 puts the algorithm into equilibrium, and thus suggest that the optimal value is 4, so the user of the algorithm does not really need to change it.

The maximum number of connections allowed for a robust cluster ($nc_r$) is also an integer equal to or greater than 1. It needs to be adjusted along with $ns_r$, because if $ns_r$ decreases, so does the number of robust clusters and thus increases the number of connections required per robust cluster. If the distance thresholds are set correctly, the accuracy should not be affected by increasing the number of allowed connections. However, this parameter may have a small impact on the processing time by limiting the number of circular connections between clusters of the same identity. We have empirically determined that the appropriate range of values for this parameter is between 5 and 25 and that it can be tuned with a sensitivity of 5.

These parameters, and especially the distance thresholds, depend mainly on the face recognition model, so they only need to be readjusted if the model is replaced. Although they may also have a small dependency on the employed dataset, it has a limited impact on the results, as we report in the subsequent ablation study.

The steps followed to fine tune the parameters are:
\begin{itemize}
    \item Iterative grid-search for tuning the three distance thresholds. The grid size starts with a size of 0.1 and is halved on each iteration until it reaches a resolution of 0.0125 (4 iterations). The total number of tests is limited according to the restrictions: $thr_{wc}$ must be smaller than $thr_f$ and that $thr_f$ must be smaller than $thr_{sc}$. In this step, $ns_r$ and $nc_r$ take fixed values of 4 and 10 respectively.
    \item Single grid-search for tuning $ns_r$ and $nc_r$. For $ns_r$ use a grid size of 1 between 3 and 6 and for $nc_r$ a grid size of 5 between 5 and 25 (20 cells in total). In this step the distance thresholds are fixed and take the values computed in the previous step.
\end{itemize}

Exploring the parameter space with the suggested approach and testing the response of the algorithm in accuracy and processing time against a training dataset, the user can easily configure the algorithm according to the application requirements. Furthermore, we parallelize every grid search in order to reduce the tuning time.

Finally, the dataset selected for tuning the parameters is reported in the description of each experiment.

\subsection{Face Clustering Performance}

For the first experiment, we use the IJB-B dataset \cite{Whitelam2017}, a well-known dataset of unconstrained in-the-wild face images. This dataset includes a clustering protocol consisting of seven subtasks that vary in the number of identities and the number of faces. We select the last subtask, as it is the most challenging one, with the highest number of identities (1,845) and faces (68,195).

For a fair comparison with other methods, and to demonstrate the OGMC algorithm is independent of the recognition model, we use the same vectors as in \cite{GCN2019} for the experiment. These vectors have 512 dimensions. For tuning the parameters of OGMC we use the provided features and labels from the CASIA dataset \cite{casia}: $thr_f$ = 1.01, $thr_{wc}$ = 1.12, $thr_{sc}$ = 0.99, $ns_r$ = 5, $nc_r$ = 5.

The performance is measured following the recommendations in \cite{metrics}, selecting the following metrics:
\begin{itemize}
    \item BCubed F-Measure $F$: represents the clustering system effectiveness, taking the bcubed precision $P$ and recall $R$ into account, which are computed as described in \cite{metrics}.
\begin{equation}
F = 2\frac{P*R}{P+R}
\end{equation}

    \item Normalized Mutual Information (NMI): this measure represents the homogeneity of the clusters. Using the ground truth clusters (G) and the predicted clusters (C) it can be computed with the following equation:
\begin{equation}
\textit{NMI}(G,C) = \frac{I(G,C)}{\sqrt{H(G)H(C)}}
\end{equation}
where $H$ represents the entropy and $I$ is the mutual information.
    \item Total processing time: since we are comparing our online algorithm with other offline methods, we process all the samples sequentially to simulate an offline behaviour and we compute the time for the whole process.
\end{itemize}

The results of the experiment are presented in Table \ref{tab:ijbb_results}. It can be observed that the proposed method outperforms the others in terms of F-Measure and processing time, while achieving competitive results in the cluster homogeneity measure. Compared to the second best method (GCN-A \cite{GCN2019}), OGMC achieves a better F-Measure while reducing the processing time by more than 6 times using the same hardware. 

\begin{table}
\caption{Comparison with baseline methods in terms of BCubed F-Measure, Normalized Mutual Information (NMI) and processing time using IJB-B 1845 subtask. Superscript* denotes results reported from the original papers, otherwise all methods use the f-vectors from \cite{GCN2019}. Superscript$^T$ denotes times reported from \cite{DDC}.}
\label{tab:ijbb_results}
\begin{center}
 \begin{tabular}{|c||c|c|c|}
 \hline
 Method & F-Measure & NMI & Run-Time\\
 \hline\hline
 K-Means$^T$ \cite{KMeans} & 0.600 & 0.868 & 00:01:00 \\ 
 \hline
 Spectral$^T$ \cite{Spectral} & 0.516 & 0.785 & - \\
 \hline
 AHC$^T$ \cite{AHC} & 0.793 & 0.923 & 00:01:32 \\
 \hline
 AP$^T$ \cite{Frey2007} & 0.477 & 0.869 & 08:42:50 \\
 \hline
 DBSCAN$^T$ \cite{DBSCAN} & 0.695 & 0.814 & 00:49:31 \\
 \hline
 \hline
 ARO$^T$ \cite{Otto2018} & 0.755 & 0.913 & 00:01:13 \\
 \hline
 PAHC*$^T$ \cite{DDC} & 0.610 & 0.890 & 00:03:56 \\
 \hline
 ConPaC*$^T$ \cite{Shi2018} & 0.634 & - & 02:53:58 \\
 \hline
 DDC$^T$ \cite{DDC} & 0.800 & 0.929 & 00:05:32 \\
 \hline
 GCN-A \cite{GCN2019} & 0.814 & \textbf{0.938} & 00:06:03 \\
 \hline
 \hline
 \textbf{OGMC (ours)} & \textbf{0.822} & 0.921 & \textbf{00:00:55} \\
 \hline
\end{tabular}
\end{center}
\end{table}

\begin{figure*}
    \centering
    \includegraphics[width=\linewidth]{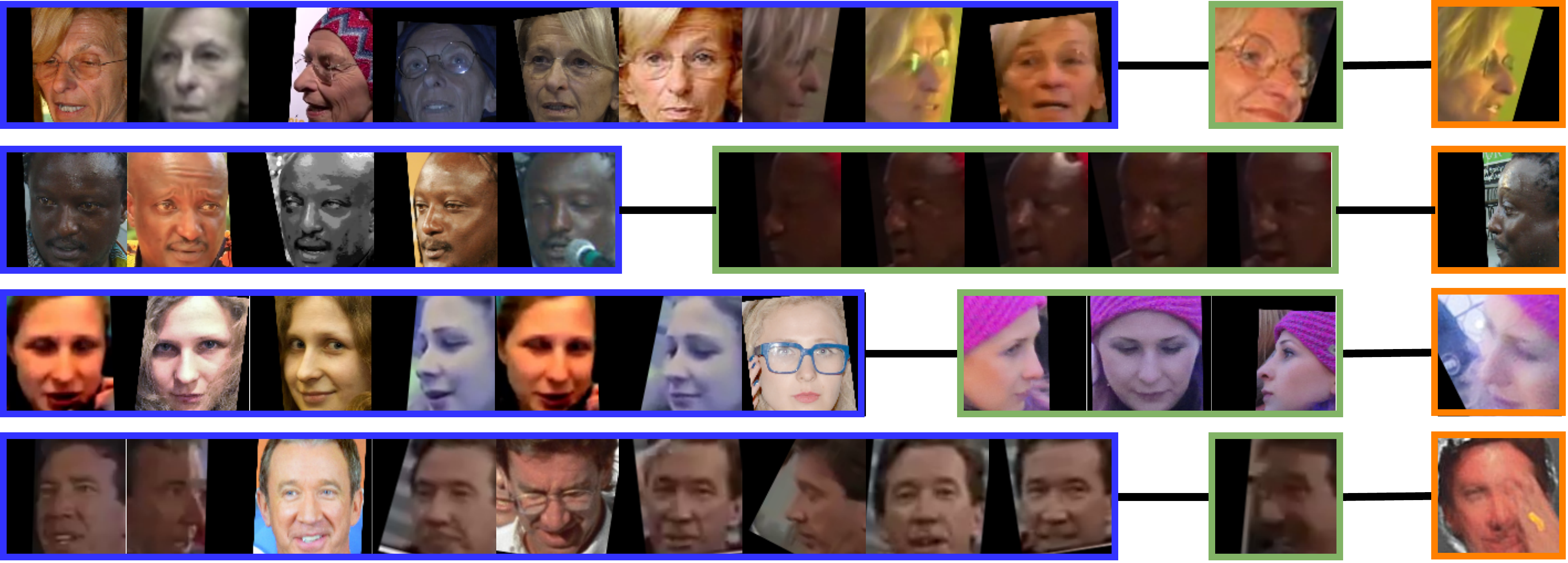}
    \caption{Example clusters and connections generated by the proposed method in the IJB-C experiment. Each row represents an identity composed of several clusters connected together.}
    \label{fig:example_connections}
\end{figure*}

In the second experiment we test the performance of our method using a different face recognition model and a different face dataset. The employed face recognition model is trained using ArcFace loss \cite{Deng2018}, with ResNet100 \cite{resnet,He2016} as the embedding network and an input resolution of $112\times112$. We select MS1MV2 \cite{Deng2018} as the training dataset, which is a refinement of MS-Celeb-1M \cite{msceleb}.

The dataset selected for this experiment is IJB-C \cite{Maze2018}, another well-known dataset of unconstrained in-the-wild face images, with a higher number of identities and faces. This dataset also includes a clustering protocol with 8 subtasks. Again, we select the most challenging protocol (IJB-C-3531), with 3,531 identities and 140,623 faces. We use RetinaFace \cite{retinaface} for the face and facial landmarks detection. In order to obtain better quality feature vectors, we filter faces with less than 45 pixels per side and we normalize the face patches applying an affine transformation using reference facial landmarks, as recommendend in \cite{Wang2018}. After filtering, 120,661 vectors belonging to 3,529 identities are extracted.

We compare our algorithm with the two best state-of-the-art methods: GCN \cite{GCN2019}, which achieved the highest accuracy in the first experiment (without considering ours), and ARO \cite{Otto2018}, which achieved the best trade-off between accuracy and speed (without considering ours). We adjust the parameters of both methods to achieve the best performance. Furthermore, for GCN, we retrain the network using a subset of VGG2 dataset \cite{vgg2}, containing over 300k images and 8500 identities, during 4 epochs (following the recommendations in \cite{GCN2019}). Finally, we readjust the parameters of our algorithm, as we are using a different face recognition model, using the same subset of VGG2 employed for retraining GCN: $thr_f$ = 1.07, $thr_{wc}$ = 1.15, $thr_{sc}$ = 1.05, $ns_r$ = 5, $nc_r$ = 10.

The results of this experiment, presented in Table \ref{tab:ijbc_results}, show that our method outperform the others in terms of F-Measure and processing time. OGMC runs 7 times faster than GCN and more than 6 times faster than ARO. These ratios of processing time show that our method is also more suitable to scale-up compared to the others.

\begin{table}
\caption{Comparison with baseline methods in terms of BCubed F-Measure and processing time using IJB-C feature vectors. All methods use the same vectors and hardware.}
\label{tab:ijbc_results}
\begin{center}
 \begin{tabular}{|c||c|c|}
 \hline
 Method & F-Measure & Run-Time\\
 \hline\hline
 ARO \cite{Otto2018} & 0.768 & 00:09:39 \\
 \hline
 GCN \cite{GCN2019} & 0.906 & 00:10:32 \\
 \hline
 \hline
 \textbf{OGMC (ours)} & \textbf{0.948} & \textbf{00:01:32} \\
 \hline
\end{tabular}
\end{center}
\end{table}

We also present several example clusters generated by the proposed method during the experiment. In Figure \ref{fig:example_connections}, each row represents an identity composed of different clusters connected to each other. These identities contain variations in lighting conditions, partial occlusions, perspective and facial attributes, so they are represented by complex data distributions. With our proposed method, we are able to accurately approximate these distributions using a variable number of connected clusters. Each robust cluster generates a centroid (representative f-vector) which represents the identity under certain condition range. For instance, in the second row of Figure \ref{fig:example_connections}, the second cluster centroid represents the identity under dim lighting conditions. If we tried to group all faces of this identity in just one cluster, it could lead to errors due to overly permissive thresholds and worse quality centroids. Another benefit of OGMC that can be observed in this figure is that the identity outliers are contained in non-robust clusters connected to the robust ones. This way, the quality of the robust cluster centroids is not compromised by these outliers and the number of matching errors are reduced. A clear example of this behaviour is exposed in the last row of Figure \ref{fig:example_connections}, where the second and the third clusters are outliers generated by a combination of unsuitable conditions (lighting, blurring, occlusion...).

In the last experiment of this section we test the accuracy of OGMC under extreme conditions: using a face recognition model which extracts vectors with less features and using a database with a much higher number of identities. Again, as we want a fair comparison with other methods we decide to replicate the experiment proposed in \cite{GCNV}. In this experiment, they selected a subset of the database MS-Celeb-1M \cite{msceleb} that contains 5.8M images from 86K identities and randomly split it into 10 parts with an almost equal number of identities. Then, they randomly selected 1 part as labeled data for training and the other 9 parts as unlabeled data. With the unlabeled data, they created 5 tests with an increasing number of vectors and identities. The last test has 5.2M vectors and 77K identities. Furthermore the provided vectors have only 256 features, compared to the 512 previously used.

Thus, as they did with the rest of the methods, we tune the parameters of OGMC using the provided training data: $thr_f$ = 0.85, $thr_{wc}$ = 1.02, $thr_{sc}$ = 0.72, $ns_r$ = 4, $nc_r$ = 25. Then, we evaluate the algorithm using the 5 test subsets with an increasing scale of vectors and identities. The results of the experiment are presented in Table \ref{tab:msceleb_results}. It can be observed that OGMC outperforms the rest of the methods consistently in every test subset.

With this experiment, we also prove that the parameters of OGMC do not need to be readjusted if the scale of the dataset increases, so the user just need to tune the parameters if the face recognition model is changed, as discussed above. Of course, as for any other method, we must ensure the training data is large and varied enough to extract the necessary information from the model to correctly tune the parameters.

\begin{table}
\caption{Comparison with baseline methods in terms of BCubed F-Measure using subsets of different sizes from MS-Celeb-1M dataset. All methods use the same 256-dimensional vectors provided by \cite{GCNV}.}
\label{tab:msceleb_results}
\begin{center}
 \begin{tabular}{|c||c|c|c|c|c|}
 \hline
 \multirow{6}{*}{Method} & \multicolumn{5}{c|}{Test} \\
    \cline{2-6}
    & \multicolumn{5}{c|}{Number of samples} \\
    \cline{2-6}
    & 584K & 1.74M & 2.89M & 4.05M & 5.21M \\
    \cline{2-6}
    & \multicolumn{5}{c|}{Number of identities} \\
    \cline{2-6}
    & 8.5K & 25.7K & 42.8K & 60.0K & 77.1K \\
    \cline{2-6}
    & \multicolumn{5}{c|}{BCubed F-Measure} \\
 \hline\hline
 K-means \cite{KMeans,mini_batch_kmeans} & 0.812 & 0.752 & 0.723 & 0.706 & 0.694 \\
 \hline
 HAC \cite{HAC} & 0.705 & 0.695 & 0.686 & 0.677 & 0.670 \\
 \hline
 DBSCAN \cite{DBSCAN} & 0.672 & 0.665 & 0.663 & 0.449 & 0.447 \\
 \hline
 ARO \cite{Otto2018} & 0.170 & 0.124 & 0.110 & 0.105 & 0.100 \\
 \hline
 CDP \cite{CDP} & 0.787 & 0.758 & 0.746 & 0.736 & 0.729 \\
 \hline
 GCN \cite{GCN2019} & 0.844 & 0.816 & 0.801 & 0.793 & 0.786 \\
 \hline
 LTC \cite{LTC} & 0.855 & 0.830 & 0.811 & 0.798 & 0.789 \\
 \hline
 GCN-V \cite{GCNV} & 0.858 & 0.826 & 0.811 & 0.799 & 0.791 \\
 \hline
 GCN-(V+E) \cite{GCNV} & 0.861 & 0.828 & 0.812 & 0.801 & 0.793 \\
 \hline
 \hline
 \textbf{OGMC (ours)} & \textbf{0.906} & \textbf{0.881} & \textbf{0.864} & \textbf{0.851} & \textbf{0.839} \\
 \hline
\end{tabular}
\end{center}
\end{table}

\subsection{Clustering Scalability}

We also want to prove that our method is scalable and that it is suitable for large-scale real-time applications. For this reason, we conduct the following two experiments. In the first one, we test the scalability of our method adding 1, 2 and 3 millions of distractors to the IJB-C vectors used in the previous experiment. Thus, we can observe how the accuracy and the processing time evolve with the size of the database. We generate the distractors from the VGG2 face dataset \cite{vgg2}, which contains 3.3 millions faces belonging to more than 9000 identities, with large variations in pose, age, illumination, ethnicity and profession. The same face recognition model of the previous IJB-C experiment is used, so we use the same parameter values: $thr_f$ = 1.04, $thr_{wc}$ = 1.13, $thr_{sc}$ = 1.03, $ns_r$ = 5, $nc_r$ = 5.

\begin{figure*}
    \centering
    \includegraphics[width=\linewidth]{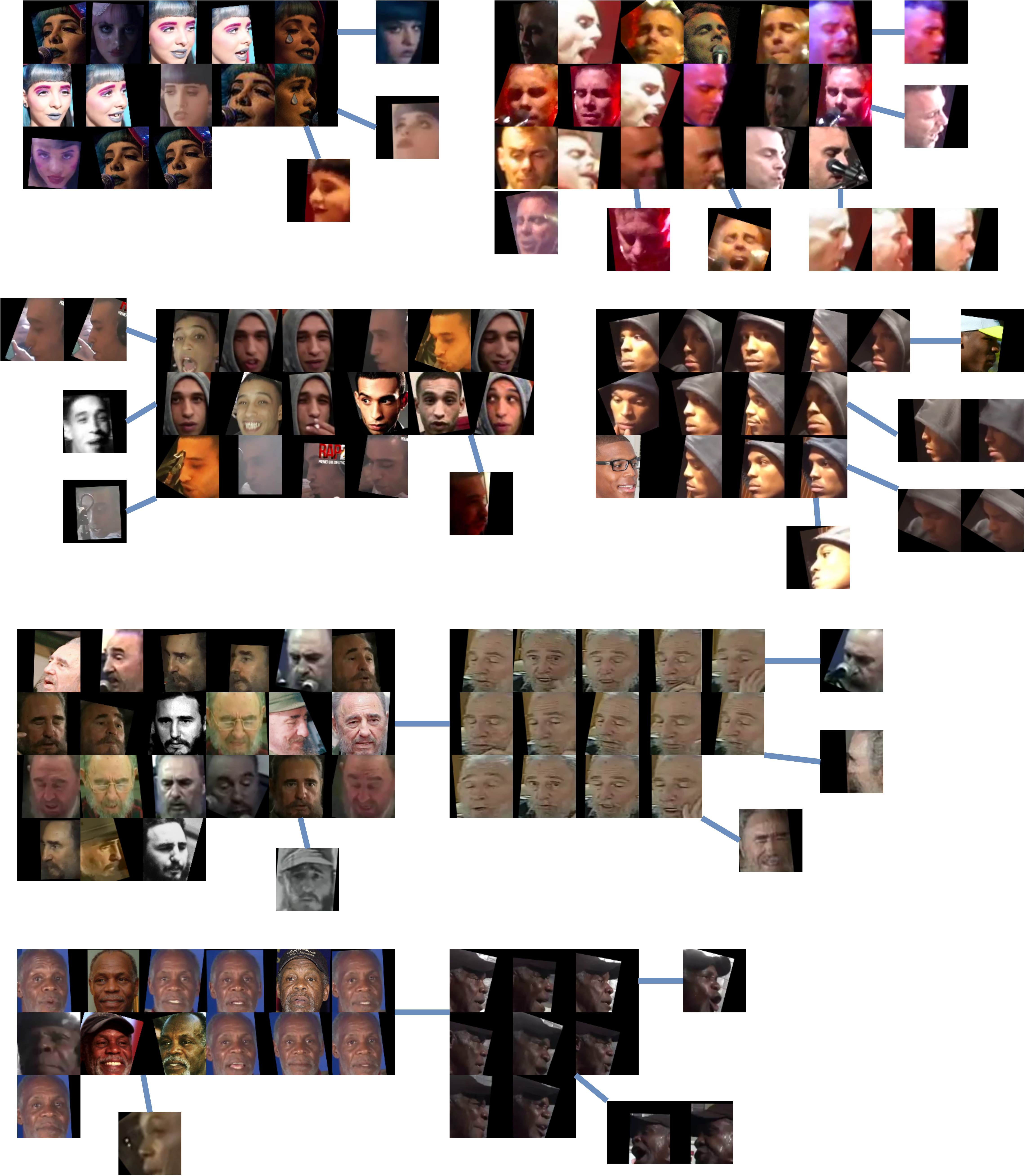}
    \caption{Example clusters and connections generated by the proposed method in the IJB-C + 3 millions of distractors from VGG2 experiment. Each group of contiguous faces represents a cluster and each line represents a connection.}
    \label{fig:cluster_examples}
\end{figure*}

Tests are executed 10 times, shuffling all the vectors a random number of times and computing the average F-Measure and processing time for all tests. For the computation of the F-Measure the distractors are ignored. The results shown in Table \ref{tab:distractor_results}, demonstrate that OGMC has an extremelly high scalability, with a drop in accuracy of 0.1\% when adding 1 million of distractors and 0.5\% when adding 3 millions. Furthermore, OGMC able to process more than 1.1 million faces in less than 25 minutes and more than 3.1 million faces in 2 hours and 13 minutes. Some examples of the clusters and connections generated in this experiment are shown in Figure \ref{fig:cluster_examples}.

\begin{table}
\caption{Results of the proposed method on IJB-C experiment adding distractors from VGG2 dataset.}
\label{tab:distractor_results}
\begin{center}
 \begin{tabular}{|c||c|c|}
 \hline
 Samples Number & F-Measure & Run-Time\\
 \hline\hline
 IJB-C & 0.952 & 00:01:32 \\
 \hline
 IJB-C + 1 million & 0.951 & 00:24:55 \\
 \hline
 IJB-C + 2 millions & 0.948 & 01:00:40 \\
 \hline
 IJB-C + 3 millions & 0.947 & 02:13:02 \\
 \hline
\end{tabular}
\end{center}
\end{table}

The last experiment aims to demonstrate that OGMC is suitable for real-time applications. It consist of repeating the previous test using IJB-C vectors with 3 millions of distractors and measuring how the processing time per sample evolves with the number of processed samples and with the number of clusters in the database (note that the number of clusters is not equal to the number of identities, as we are not taking connections into account).

\begin{figure}
    \centering
    \includegraphics[width=\linewidth]{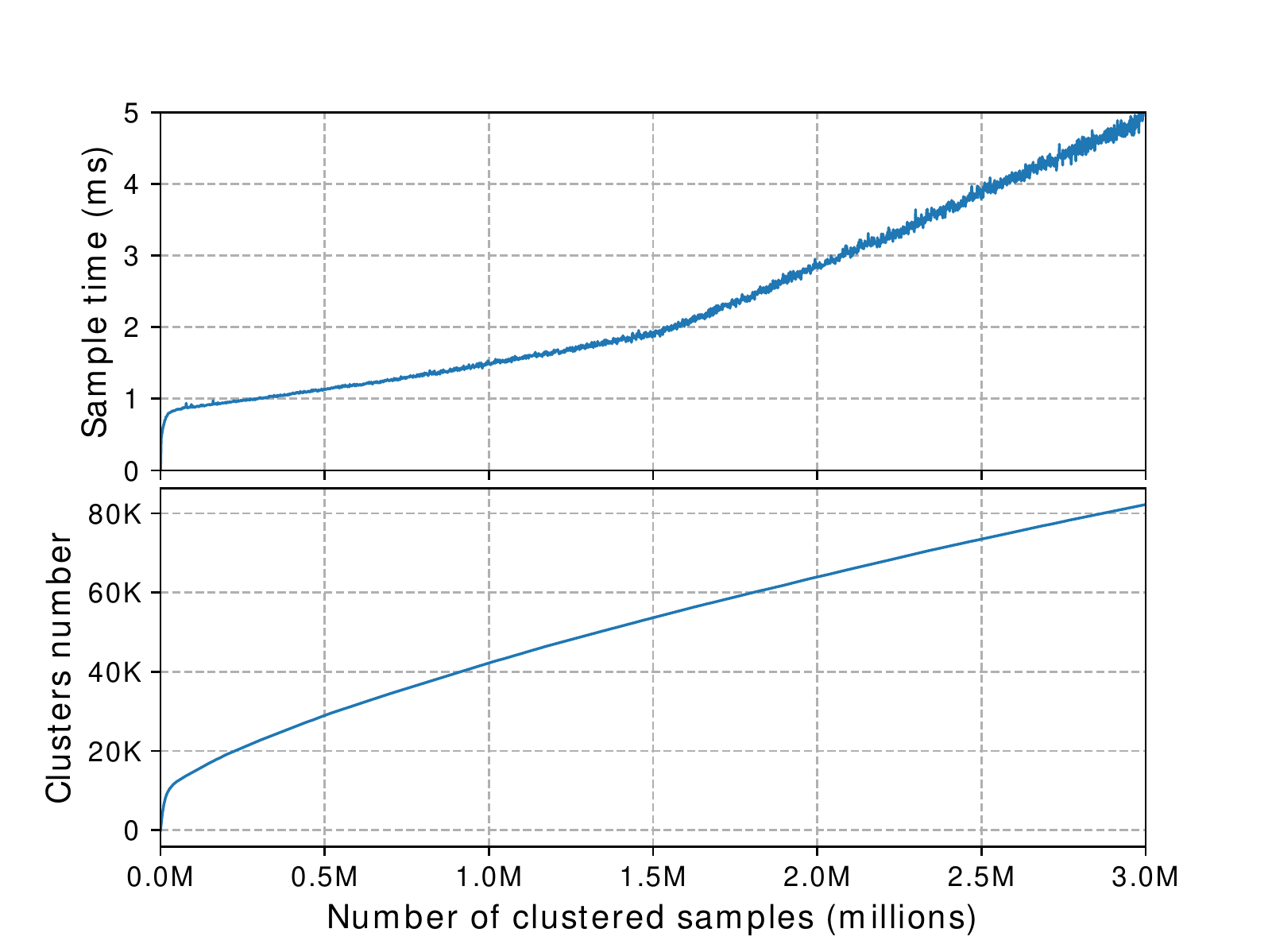}
    \caption{Evolution of the clustering time per sample with the number of samples clusterized compared to the evolution of the number of clusters in the database.}
    \label{fig:clustering_times}
\end{figure}

As in the previous experiment, the test is repeated 10 times, shuffling all the vectors a random number of times and computing the results as the average value of all tests. Figure \ref{fig:clustering_times} shows that, after processing 3 millions samples and with a database of 80 thousand clusters, the processing time per sample is still 5 milliseconds, which still allows OGMC to process 200 samples per second in real-time. This results demonstrate that the proposed method is suitable for real-time applications, even when dealing with extremelly large amounts of data.

\subsection{Ablation Study}

We present an ablation study to discuss about the impact of design choices on the performance of OGMC. First, we demonstrate the contribution of the reclustering stage of the algorithm. For this purpose, we run the IJB-C with VGG2 distractors experiment applying only the first stage of OGMC and compare the results with the ones obtained applying the full method. Table \ref{tab:two_stages} shows the results of the experiment. Comparing the processing times, we can see that removing the reclustering stage significantly reduces the complexity of the algorithm. However, it also leads to an important drop in accuracy ($\approx$4\%). Furthermore, removing the second stage also significantly increases the dependency of accuracy on the order of the input data. This is because the connections are not re-evaluated when the clusters are updated.

\begin{table}
\caption{Comparison of the first stage of OGMC against the full method.}
\label{tab:two_stages}
\begin{center}
 \begin{tabular}{|c||c|c|c|c|}
 \hline
 \multirow{2}{*}{Dataset} & \multicolumn{2}{c|}{Full method} & \multicolumn{2}{c|}{First Stage} \\
    \cline{2-5}
     & F-Meas & Run-Time & F-Meas & Run-Time \\
 \hline\hline
  IJB-C & 0.952 & 00:01:32 & 0.914 & 00:00:49 \\
 \hline
 IJB-C + 1 million & 0.951 & 00:24:55 & 0.912 & 00:10:12 \\
 \hline
 IJB-C + 2 millions & 0.948 & 01:00:40 & 0.907 & 00:23:11 \\
 \hline
 IJB-C + 3 millions & 0.947 & 02:13:02 & 0.901 & 00:59:18 \\
 \hline
\end{tabular}
\end{center}
\end{table}

\begin{figure}
    \centering
    \includegraphics[width=\linewidth]{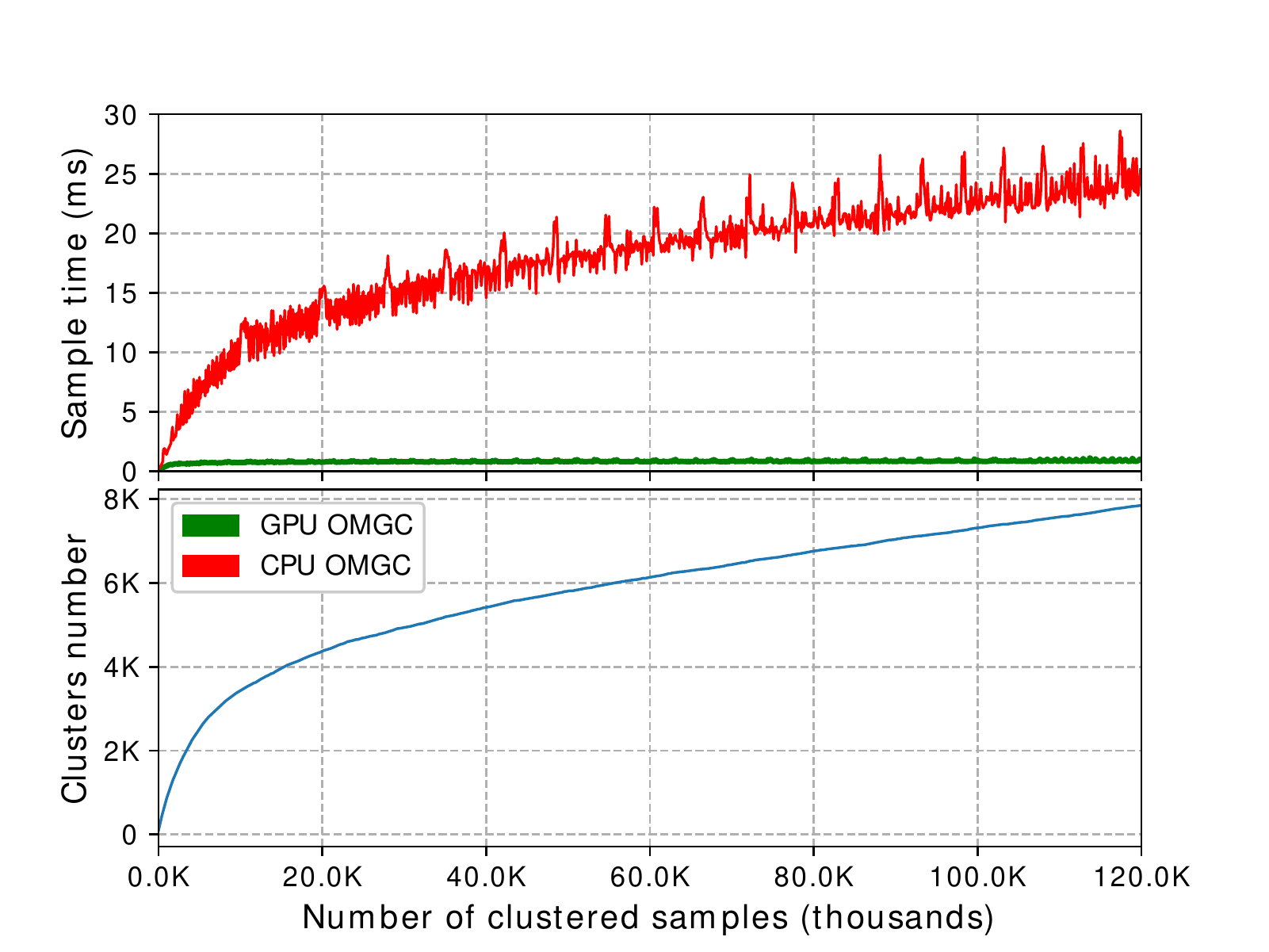}
    \caption{Comparison of the evolution of the clustering time per sample with the number of samples clusterized for GPU and CPU versions of OGMC.}
    \label{fig:gpu_comp}
\end{figure}

We also want to analyze the speed up added by the GPU parallelization. We reimplement the distance computing module with CPU and repeat the IJB-C experiment with both versions of OGMC to measure how the processing time evolves with the number of clusterized samples in each case. The results of the comparison in Figure \ref{fig:gpu_comp} show that using the GPU for the distance computing parallelization produces a huge decrease on the computation time per sample and a huge increase on the scalability. However, this simple task only consumes a small percentage of the GPU utilization compared to other methods based on neural networks as GCN \cite{GCN2019} or GCN-V \cite{GCNV}. Furthermore the GPU memory needed for the cluster centroids is very limited (2.4GB for 1 million 512-dimensional vectors) and can be allocated dynamically as the database growths, which makes OGMC suitable to run together with other GPU-based algorithms (for example a DNN-based face recognition model).

Finally, we want to demonstrate that the parameters of OGMC do not have a significant dependency on the training dataset, as long as it is large and varied enough to extract the necessary information from the face recognition model. In order to prove this, we repeat the IJB-C experiment with the same face recognition model but, instead of using the VGG2 subset, we tune the parameters using the IJB-C vectors, so that we obtain the best possible results in this test. Thus, we obtain the following values for the parameters: $thr_f$ = 1.04, $thr_{wc}$ = 1.13, $thr_{sc}$ = 1.03, $ns_r$ = 5, $nc_r$ = 5. With these values OGMC achieves a BCubed F-Measure of 0.952, only a 0.4\% of improvement over the original test. Thus, this experiment suggests our hypothesis is correct, the OGMC parameters depend almost entirely on the face recognition model and they only need to be readjusted if such model is replaced.

\subsection{Beyond Face Recogniton}

To conclude the experimental section, we evaluate the effectiveness of OGMC for tasks beyond face recognition. For this purpose, we reproduce the experiment presented in \cite{GCNV} with DeepFashion dataset \cite{deepfashion}, a well-known dataset for clothes retrieval. In this experiment they mix the training and testing features in the original split, and randomly sample 25,752 images from 3,997 categories for training and the other 26,960 images with 3,984 categories for testing. For a fair comparison, we use the same 256-dimensional vectors provided in their official repository. Thus, we tune the parameters of OGMC using the training set: $thr_f$ = 0.51, $thr_{wc}$ = 0.58, $thr_{sc}$ = 0.49, $ns_r$ = 4, $nc_r$ = 10. The results of the experiment are presented in Table \ref{tab:deepfashion_results}. Among all the tested methods, OGMC achieves the best F-Measure, demonstrating its suitability for tasks beyond face recognition.

\begin{table}
\caption{Comparison with baseline methods in terms of BCubed F-Measure with DeepFashion dataset. All methods use the same vectors from \cite{GCNV}.}
\label{tab:deepfashion_results}
\begin{center}
 \begin{tabular}{|c||c|c|c|}
 \hline
 Method & F-Measure \\
 \hline\hline
 K-Means \cite{KMeans,mini_batch_kmeans} & 0.538 \\ 
 \hline
 HAC \cite{HAC} & 0.488 \\ 
 \hline
 DBSCAN \cite{DBSCAN} & 0.532 \\ 
 \hline
 MeanShift \cite{meanshift,meanshift1} & 0.567 \\ 
 \hline
 Spectral \cite{spectral1,spectral2} & 0.464 \\ 
 \hline
 ARO \cite{Otto2018} & 0.530 \\ 
 \hline
 CDP \cite{CDP} & 0.578 \\ 
 \hline
 GCN \cite{GCN2019} & 0.589 \\ 
 \hline
 LTC \cite{LTC} & 0.591 \\ 
 \hline
 GCN-V \cite{GCNV} & 0.573 \\ 
 \hline
 GCN-(V+E) \cite{GCNV} & 0.601 \\
 \hline
 \hline
 \textbf{OGMC (ours)} & \textbf{0.620} \\
 \hline
\end{tabular}
\end{center}
\end{table}

\section{Conclusions}

In this work, we address the problem of large-scale online face clustering. We propose a method based on EM algorithm and Mixture of Gaussians. We introduce the concept of cluster connection, where an identity can be represented by multiple clusters. With this approach, we reduce the dependency of the clustering process on the order and the size of the incoming data and we are able to deal with complex data distributions. The conducted experiments and their derived results show that our method outperforms the state-of-the-art clustering methods, regardless of whether they are online or offline, in terms of accuracy, processing time and scalability.

Future work will focus on further reducing the processing time of the proposed method by also parallelizing the minimum distance search using the computational capabilities of the GPU. In addition, given the current situation with Covid-19, we will study the impact of the use of surgical masks on the clustering process. We believe that for an identity, the proposed method would group faces with and without masks in different clusters, but connected to each other, since we have proved that the algorithm works well with partial occlusions. Finally, as we believe in the open source community, we will soon release the full code of OGMC and all the supplementary material used for the experiments, so anyone can replicate them and contribute to improve the method.

\bibliographystyle{IEEEtran}
\bibliography{biblio}

\begin{biography}[{\includegraphics[width=1in,clip,keepaspectratio]{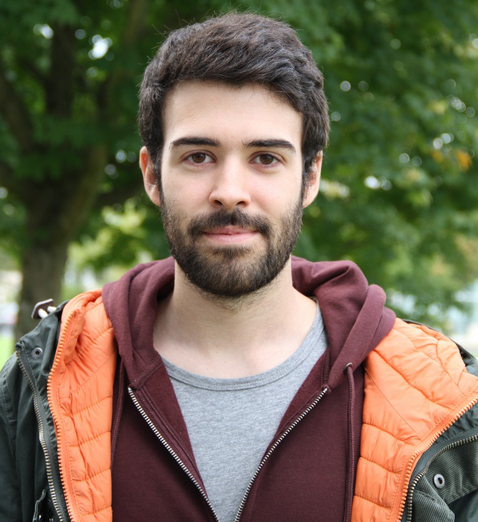}}]{David Montero}
 received his B.Sc. degree in Industrial Engineering and his M.S. degree in Industrial Engineering from ETSI of the University of Seville, Spain, in 2015 and 2017 respectively. Since 2017 he has worked as a Researcher on computer vision and machine learning techniques of the Intelligent Transport Systems and Engineering Area of Vicomtech. Since 2019 he is also working towards the Ph.D. degree with the Department of Computer Science and Artificial Intelligence of the University of the Basque Country. His research interests include pattern recognition and computer vision.
\end{biography}
\vskip 0pt plus -1fil
\begin{biography}[{\includegraphics[width=1in,clip,keepaspectratio]{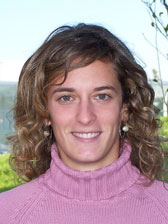}}]{Naiara Aginako}
 received her degree in Telecommunications Engineering from the University of the Basque Country (UPV/EHU) in 2005, and her Ph.D. degree in computer sciences from the University of the Basque Country in 2017. From 2003 to 2005 she collaborated with the Signal Processing and Communication Group of the Department of Electronics and Telecommunications. She worked as a senior researcher in Vicomtech from 2005 until 2015, in the Digital Media department. From 2015 until 2018 she taught in the Engineering School of Gipuzkoa in the Applied Mathematics department. Since 2018 she has been working in the Faculty of Informatics in the Computer Science and Artificial Intelligence department (UPV/EHU). Her research is mainly based on Computer Vision more specifically on Image and Video Analysis. Her work as a researcher includes a series of publications and two patents.

\end{biography}
\vskip 0pt plus -1fil
\begin{biography}[{\includegraphics[width=1in,clip,keepaspectratio]{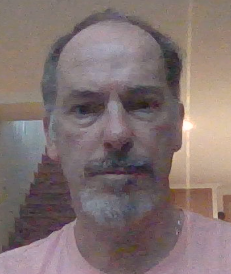}}]{Basilio Sierra}
 received the B.Sc. degree in computer sciences, the M.Sc. degree in computer science and architecture, and the Ph.D. degree in computer sciences from the University of the Basque Country, in 1990, 1992, and 2000, respectively. He is currently a Full Professor with the Computer Sciences and Artificial Intelligence Department, University of the Basque Country. He is also the Co-Director of the Robotics and Autonomous Systems Group, Donostia-San Sebastian. He is also a Researcher in the fields of robotics and machine learning, where he is working on the use of different paradigms to improve behaviours. He has published more than 50 journal articles, and several book chapters and conference papers.
\end{biography}
\vskip 0pt plus -1fil
\begin{biography}[{\includegraphics[width=1in,clip,keepaspectratio]{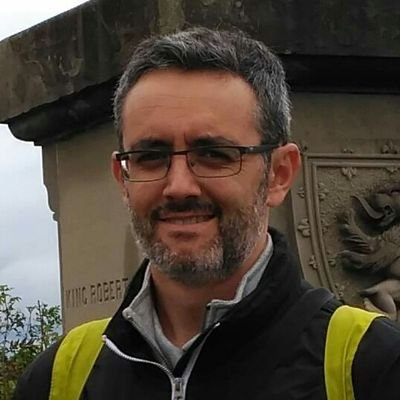}}]{Marcos Nieto}
 received his M.S. and Ph.D. degree in Electrical Engineering from ETSIT of the Universidad Politécnica de Madrid (UPM), Spain, in 2005 and 2010, respectively. From 2005 to 2010 he worked as Researcher within the Image Processing Group at the UPM. Since 2010 he has worked as a Researcher on computer vision and machine learning techniques of the Intelligent Transport Systems and Engineering Area of Vicomtech. He has been the technical and scientific coordinator of FP7 and H2020 projects and has experience in transferring technology to industry to support innovation. He is the author of more than 60 peer reviewed international publications in relevant conferences (+40) and journals (+15). He is also an active reviewer of prestigious journals for IEEE, Elsevier and Springer.
\end{biography}

\end{document}